\documentclass[acmlarge,screen]{acmart}
\AtBeginDocument{%
  \providecommand\BibTeX{{%
    \normalfont B\kern-0.5em{\scshape i\kern-0.25em b}\kern-0.8em\TeX}}}

\setcopyright{acmlicensed}
\copyrightyear{2024}
\acmYear{2024}
\acmDOI{XXXXXXX.XXXXXXX}

\acmConference[ICISDM '24]{8th International Conference on Information System and Data Mining}{June 24th-26th, 2024}{Los Angeles, CA}
\usepackage{amsmath}
\usepackage{tabularx}
\usepackage{hyperref}
\usepackage{booktabs}
\usepackage{multirow}
\begin{document}

\title{Relations Prediction for Knowledge Graph Completion using Large Language Models}

\author{Sakher Khalil Alqaaidi}
\email{sakher.a@uga.edu}
\author{Krzysztof Kochut}
\email{kkochut@uga.edu}
\affiliation{%
  \institution{University of Georgia}
  \city{Athens}
  \state{Georgia}
  \country{USA}
}

\renewcommand{\shortauthors}{Alqaaidi and Kochut, et al.}

\begin{abstract}
Knowledge Graphs have been widely used to represent facts in a structured format. Due to their large scale applications, knowledge graphs suffer from being incomplete. The relation prediction task obtains knowledge graph completion by assigning one or more possible relations to each pair of nodes. In this work, we make use of the knowledge graph node names to fine-tune a large language model for the relation prediction task. By utilizing the node names only we enable our model to operate sufficiently in the inductive settings. Our experiments show that we accomplish new scores on a widely used knowledge graph benchmark.
\end{abstract}





\maketitle

\section{Introduction}
Knowledge Graphs (KGs) are used to store semantic data in the form of entity nodes and edges. The edges represent the directed relations between the entities. For instance, the fact that \textit{James Cameron produced Avatar} can be stored in a KG by having \textit{James Cameron} node linked to \textit{Avatar} node using the relation (edge) \textit{produced}. This knowledge representation structure is called a triple; it consists of a head node, a relation, and a tail node. The direction in this triple is important to keep the validity of the fact if the nodes that surround the edge are swapped. Due to KGs efficiency in representing semantic facts, KGs are used in large scale applications, such as recommender systems and information extraction \cite{wang2017knowledge}. However, due to their enormous sizes, KGs suffer from being incomplete \cite{west2014knowledge}. Consequently, several methods were proposed for Knowledge Graph Completion (KGC) \cite{shen2022comprehensive}. Predicting the relations between node pairs is one of these methods. The Relation Prediction (RP) task aims at identifying the relations between two given nodes; the nodes order in the RP task input is important to maintain the relation direction and to distinguish between the head node from the tail node. Formally, in the RP task, a function $f$ is trained to predict a set of relations $R$ for a given nodes pair, that are the head node $n_h$ and the tail node $n_t$, as in the following: $f(n_h,n_t) = R$. Therefore, in the previously given example, the RP task could also predict additional plausible relations between \textit{James Cameron} node and \textit{Avatar} node, such as \textit{writer\_for}, \textit{directed}, \textit{honored\_for}, and \textit{nominated\_for}.

Recent achievements in KGC were obtained by encoding nodes and edges into numerical float vectors \cite{shen2022comprehensive}. These vectors are commonly called embeddings and they are widely used in machine learning-based models. Initially, KGC models relied solely on the graph structure to generate embeddings for nodes and edges \cite{bordes2013translating,sun2019rotate}. The graph structural information reflects the nodes topological details, such as the node degree, the lengths of the walks that can be started from the node, and the structure of the node's neighborhood. However, KGC structure-based models had unsatisfying performance in the inductive settings, that is when predicting relations for nodes not seen during the model training.

\begin{table}[t]
\centering
\caption{Main differences and features of the related work compared to our model.}
\label{tab:models}
\begin{tabular*}{\linewidth}{@{\extracolsep{\fill}}lcccc}
\hline
Model&\shortstack{Entity \\ Description-Free}& \shortstack{Inductive \\ Settings}&\shortstack{Single-Stage \\ Training}&Input Type\\
\hline
TransE \cite{bordes2013translating}&\checkmark&$\times$&$\checkmark$&Graph structure\\
PTransE \cite{lin2015modeling}&\checkmark&$\times$&$\times$&Graph structure\\
TKRL \cite{xie2016representation}&\checkmark&$\times$&$\times$&Node types and Graph structure\\
DKRL \cite{xie2016representation2}&$\times$&$\times$&$\times$&Node descriptions and Graph structure\\
KGML \cite{onuki2019relation}&\checkmark&$\times$&\checkmark&Graph structure\\
Shallom \cite{demir2021shallow}&\checkmark&$\times$&\checkmark&Graph structure\\
TaRP \cite{cui2021type}&$\checkmark$&$\times$&$\times$&Node types and structural info.\\
KG-BERT \cite{devlin2018bert}& $\times$ &\checkmark&\checkmark&Node names and descriptions\\
\hline
RPLLM (Ours)&\checkmark&\checkmark&\checkmark&Node names\\
\hline
\end{tabular*}
\end{table}

On the other hand, a stream of KGC models employed the embeddings generated from the nodes' content, such as the text in node labels or the numerical values in node attributes \cite{yao2019kg,zha2022inductive}.
However, due to the heterogeneity of the content types, several KGC models only utilized the textual content supported by the easiness of text encoding using language models. Furthermore, the ability to handle nodes' text not seen during the model training led to the superiority of language model-based KGC works in the open-world scenario.

In Natural Language Processing (NLP), the employment of language models along with machine learning has led to astonishing accomplishments on a variety of tasks, particularly, after the emerge of Pre-trained Language Models (PLMs), such as BERT, GPT2, and Roberta \cite{min2023recent}. Recently, Large Language Models (LLMs) have outperformed PLMs in several NLP tasks due to their enormous training corpora and advanced transformers \cite{touvron2023llama}.

In this work, we utilize LLMs to complete knowledge graphs. Particularly, we exploit Llama 2 \cite{touvron2023llama}, a new yet powerful LLM. We use Llama 2 to perform sequence multi-label classification for the relation prediction task and achieve a new score in the Freebase benchmark. Additionally, we achieve scores that are equivalent to the  reported ones in the literature for the WordNet benchmark; we justify these scores in Section \ref{sec:expertiments}. Our implementation works well with the inductive settings as we depend on the nodes textual information only to predict the relations. Moreover, our model is advantageous over other models because we employ the node names only, i.e., neglecting the lengthy node descriptions when fine-tuning the model.

\section{Related Work} \label{sec:related}

\subsection{Knowledge Graph Completion}
KGC models varied by the followed method, namely, though not exhaustively, rule-based methods \cite{kuvzelka2020markov,zhang2020efficient}, tensor factorization methods \cite{balavzevic2019tucker, nickel2011three,tay2017random}, structural translation-based methods \cite{bordes2013translating,lin2015modeling}, and KG content utilization methods \cite{yao2019kg,wang2021structure}. As well as that, KGC models varied by the completion objective. A stream of models targeted completing KGs by predicting missing entities in KG triples \cite{bordes2013translating, hu2019knowledge}. Entity Prediction (EP) or Link Prediction (LP) were common names in this direction. Similarly another KGC objective predicts the relations in node pairs, that is important to enriching KGs in two ways. First, through establishing missing links between nodes. Second, by fostering the already connected nodes using multi-relational links.

Here we review the related relation prediction models and highlight their main differences and features compared to our work in Table \ref{tab:models}. TransE \cite{bordes2013translating} proposed a translation method to represent nodes and edges in graphs using their structural information. The model training objective was achieving low energy score for the translation equation $E(h,r,t)=||h + r - t||$, where $E$ is the energy score, $h$, $r$ and $t$ are the head node, the edge, and the tail node embeddings, respectively. PTransE \cite{lin2015modeling} extended TransE to utilize relation paths in representation learning for KGs. Instead of considering direct edges between nodes only, PTransE took into account multi-step relation paths. TKRL \cite{xie2016representation} is a model that also follow the translation technique in TransE empowered by the node entity types for the KGC task. DKRL \cite{xie2016representation2} used TransE embeddings and aligned them with the nodes' description embeddings. Word2Vec \cite{mikolov2013efficient} was employed for the descriptions encoding, then the embeddings were fed to a convolutional neural network. KG-BERT \cite{yao2019kg} considered the triple's textual content only as the input for BERT language model, then fine-tuned the model for the LP task and the RP task. KGML \cite{onuki2019relation} is a simple yet effective RP model. KGML's authors represented the entities using one-hot encoder and used the input in a multi-layer neural network. TaRP \cite{cui2021type} incorporated the entity types in the model decision level not in the embeddings learning. The model proposed multiple variants based on different KG embeddings method, such as TransE and RotateE \cite{sun2019rotate}. Shallom \cite{demir2021shallow} trained a shallow neural network to learn nodes pair embeddings, then to predict their possible relations. Our model has advantages compared to the ones mentioned, thanks to the following features. The utilization of the minimal node information represented by the node names only, the ability to operate in the inductive settings, and having single staged training as shown in Table \ref{tab:models}.

\subsection{Language Models}

Text features extraction emerged from finding single word representation \cite{pennington2014glove} to sentences and documents representation. Recurrent Neural Networks (RNNs) \cite{schuster1997bidirectional} are capable of comprehending sequences of words, yet their effectiveness is constrained by memory limitations, restricting the scope of words consideration. Consequently, the contextualized representation of sentences in RNNs often fell short of satisfaction. Later, the attention technique \cite{vaswani2017attention} led to building the transformer architecture. Using attention transformers, NLP models became capable of grasping complete sentences or paragraphs and present an meaningful representation. In light of the enhancements in the computational technologies, more advanced transformer models were trained on large corpora with self-supervised learning objectives. These models are called Pre-trained Language Models (PLMs); some remarkable PLM examples are BERT and GPT \cite{devlin2018bert,radford2018improving}. Recently, Large Language Models (LLMs) used auto-regressive transformer architectures for pre-training on even larger corpora, that included web crawl dumps of online encyclopedias, books, academic papers and spoken dialogues in movies and interviews. Perceiving such substantial text content led to handling more complicated language rules and expert knowledge across several domains simultaneously. PLMs and LLMs were used in a variety of applications by fine-tuning them for downstream tasks. Since their emergence, LLMs have been entertaining us with significant applications, such as chatbots capable of achieving human-level performance, as demonstrated in ChatGPT. In this work, we efficiently employ LLMs in the RP task by showing its superiority over works that used other language models.


\section{Methodology}

\begin{figure}[h]
  \centering
  
  \includegraphics[width=\linewidth]{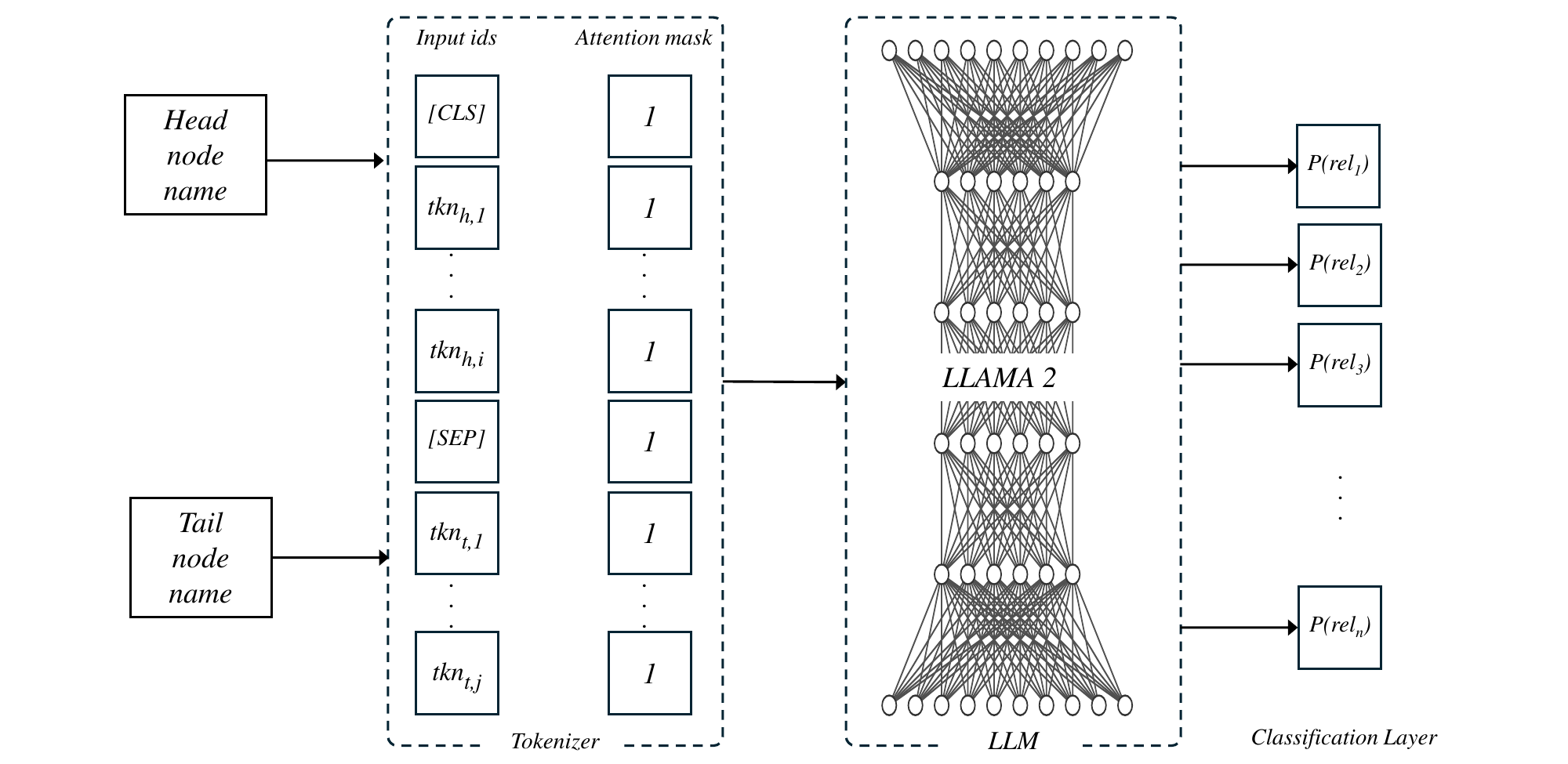}
  \caption{Our model}
  \Description{The description.}
  \label{fig:diagram}
\end{figure}

We exploit a powerful large language model, Llama 2 \cite{touvron2023llama}, to predict the plausible relations between entity pairs. The entities represent nodes in KGs and the relations set is fixed. Figure \ref{fig:diagram} shows the main architecture of our implementation. We fine-tune Llama 2 for a downstream supervised task, that is multi-label sequence classification. We validate the fine-tuning process based on the categorical cross entropy loss
\begin{align*}
\ell(\hat{y}, y) = - \sum_{r=1}^{R} log \dfrac{exp(\hat{y_r})}{ \sum^R_{i=1} exp(\hat{y_i})}.y_r
\end{align*} where $\hat{y}$ denotes the predicted relation probabilities, $y$ denotes the ground-truth binary labels, and $C$ denotes the relations size.

\subsection{Text Encoding}

To achieve a straightforward and highly effective implementation, we opt to utilize only entity names as input for the LLM. LLMs' pre-training and fine-tuning stages involve text tokenization, followed by encoding the tokens into numerical IDs \cite{song2020fast}. Accordingly, we utilize the Llama 2 tokenizer to convert the entity text into numerical IDs. The tokenizer matches each word in the text sequence with an ID in its vocabulary. If there is no match, the tokenizer breaks the word into chunks. The tokenizer uses the chunks to find new matches or keeps breaking them until reaching the character level in a recursive operation.

In addition to the input IDs, the tokenizer uses special tokens, \textit{[CLS]} and \textit{[SEP]}. The former denotes the beginning of the sequence, whereas the later is used to separate each entity word sequences. To maintain consistency in the input ID lists, padding and truncation actions are applied. Accordingly, if the size of the input IDs list is smaller than a defined padding length, a padding token is used to fill the gaps. Alternatively, if the input IDs list is larger, a truncation process is applied. Besides the input IDs list, the tokenizer returns an attention mask, that is a sequence of binary values. The items with 1 in the attention mask indicate the positions of the tokens in the input IDs list, whereas the zeros indicate padding tokens, if any.

\subsection{Llama 2}
Llama 2 receives the input IDs and the attention mask from the tokenizer, then processes these values through several layers of attention transformers. Each layer's parameters are fine-tuned based on the provided relation labels. Llama 2 for sequence classification has a top linear layer, which outputs the final relations probabilities.

\begin{table}[t]
\centering
\caption{Statistics of the experimental datasets.}
\label{tab:datasets}
\begin{tabular*}{\linewidth}{@{\extracolsep{\fill}}lccccc}
\hline
\multirow{ 2}{*}{Dataset}&\multicolumn{3}{c}{Triples}&\multirow{ 2}{*}{Entities}&\multirow{ 2}{*}{Relations}\\
\cline{2-4}
&Training&Validation&Testing&&\\
\hline
FreeBase&483,142&50,000&59,071&14,951&1,345\\
WordNet&141,442&5,000&5,000&40,943&18\\
\hline
\end{tabular*}
\end{table}

\section{Experiments} \label{sec:expertiments}

\subsection{Experimental Setup}\label{sec:experimental setup}
To showcase the effectiveness of our model, we conduct evaluations on two widely recognized benchmarks, FreeBase \cite{bollacker2008freebase} and WordNet \cite{miller1995wordnet}. Table \ref{tab:datasets} shows statistics of the two KGs. Negative sampling is a common practice in the LP task, often resulting in the duplication of training triples. Specifically, this method involves corrupting copies of the training triples by replacing the original positive label with a negative one. The corruption typically entails substituting a correct entity with an invalid, randomly selected entity. While this approach aims to enhance the diversity of training data and improve model performance, it inevitably prolongs training time. However, when it comes to the RP task, the implementation of negative sampling presents a distinct research challenge. This is primarily because the RP task involves assigning multiple labels to node pairs, unlike the binary labels used in the LP task. Consequently, we opt not to incorporate negative sampling into the RP task due to this fundamental difference in label assignment methodology.

The Llama 2 collection of pre-trained large language models varies in parameter count from 7 billion to 70 billion. Considering our computational resources, we opt to fine-tune the Llama 2 model pre-trained with 7 billion parameters for the RP task. We use Nvidia A100-SXM-80GB as the GPU node. We fine-tune Llama-2-70B using PyTorch for 10 epochs only. The padding length in Llama's tokenizer is 50 for the entities' text sequence. We use Adam algorithm for optimization. The learning rate is 5e-5 and the optimizer decay is 25\%.


\subsection{Evaluation Metrics} \label{sec:metrics}
The Mean Rank (MR) and the Hits@N are commonly used metrics for evaluating predictions ranking, particularly in KGC tasks. Given that the model predictions represent relation probabilities, the MR metric calculates the mean rank of the ground truth relation in the probabilities sorted in descending order. Formally defined as
\begin{align*}
MR = \dfrac{1}{|Q|} \sum^{|Q|}_{i=1}rank_i
\end{align*}
where $Q$ is the query set. Accordingly, KGC models aim to achieve low MR values. In this context, filtered settings  \cite{bordes2013translating} are particularly useful considering the potential for multiple relations to be predicted for a pair of nodes. Therefore, in this evaluation scenario, the prediction rank should be decreased by the number of valid relations that were ranked above the predicted one. We report the filtered settings performance in our experiment results.

The Hits@N metric finds the average number of ground truth relations found in N probabilities sorted in descending order. Formally defined as
\begin{align*}
Hits@N = \dfrac{1}{|Q|} \sum^{|Q|}_{i=1}rank\_hit(rank_i, N)
\end{align*}

\begin{align*}
rank\_hit(rank, N) = \left\{ \begin{array}{ll}
         1, & rank \leq N \\
         0, & rank > N
    \end{array}
    \right.
\end{align*}
where $Q$ is the query set. Commonly, KGC models report this metric with $N=1$. Similarly, we do so. We also use the filtered settings for this metrics in our results.

We extend the reported scores of our model to show the results on Hits@5 in Table \ref{tab:inductive}. Additionally, we show the Mean Reciprocal Rank (MRR) score in the same table. The MRR metric is similar to the MR metric, with the distinction that it represents the score as a percentage value. The following equation is a formal definition of the MRR metric.
\begin{align*}
MR = \dfrac{1}{|Q|} \sum^{|Q|}_{i=1}\dfrac{1}{rank_i}
\end{align*}

\subsection{Comparison Models}
We conduct a comparison of our model with various RP baselines. Section \ref{sec:related} briefly explains the methodology of each model used in the comparison. The RP scores in TransE \cite{bordes2013translating} were reported by several other papers and we use their scores in our comparison. PTransE \cite{lin2015modeling} presented several variants based on the embeddings concatenation method and the path length. We report their best variant's score in our comparison. TKRL \cite{xie2016representation} proposed a recursive variant for building hierarchical type projection matrices, which achieved the best performance in the RP task among the other variants. Similarly, DKRL \cite{xie2016representation2} proposed several variants. However, the variant with convolutional layers and TransE embeddings achieved the best scores. In KGML \cite{onuki2019relation}, $\alpha$ is a balancing parameter between their combined loss functions. We report their best scores that appeared in two different $\alpha$ parameters. Shallom \cite{demir2021shallow} differs from KGML by employing a concatenation step for the entity embeddings. Also, differs in the cross entropy loss utilization. TaRP \cite{cui2021type} presented three variants based on the embeddings sub-model. The best performing variant is the one that employed RotateE \cite{sun2019rotate} for KG embeddings. The KG embeddings in TaRP were used along with a relation decision level stage, that depended on the node types. We share a similar architecture with KG-BERT \cite{yao2019kg} with two differences, that are the used language model and the node information usage. Given that we were able to reproduce KG-BERT's results using the author's published code, other models' code was not available online, such as TaRP.

\begin{table}[t]
\centering
\caption{The mean rank and Hits@1 relation prediction results on FB15K, including the filtered settings.}
\label{tab:results}
\begin{tabular*}{\linewidth}{@{\extracolsep{\fill}}lcccc}
\hline
Model&Mean Rank&Filtered Mean Rank&Hits@1&Filtered Hits@1\\
\hline
TransE \cite{xie2016representation} & 2.10  & -& 0.65 & 0.84\\
PTransE (ADD, len-2 path) \cite{lin2015modeling} & 1.70  & 1.20& 0.69 & 0.93\\
TKRL (RHE) \cite{xie2016representation} & 2.12  & 1.73& 0.71 & 0.93\\
DKRL (CNN)+TransE \cite{xie2016representation2}& 2.41  & 2.03& 0.70 & 0.91 \\
KGML ($\alpha = 1.0$) \cite{onuki2019relation} & - &-& 0.73 & 0.94\\
KGML ($\alpha = 0.0$) \cite{onuki2019relation} & - &-& 0.66 & \textbf{0.96}\\
Shallom \cite{demir2021shallow}    &1.59      &-&0.73& -\\
TaRP RotateE \cite{cui2021type}    & -   &1.16&  -& 0.93\\
KG-BERT \cite{yao2019kg}   & 1.69   &1.25&0.69   & \textbf{0.96}\\
RPLLM (Ours)    &\textbf{1.50}      &\textbf{1.15}&\textbf{0.74}& 0.95\\
\hline
\end{tabular*}
\end{table}

\begin{table}[t]
\centering
\caption{The mean rank and Hits@1 relation prediction results on WordNet, including the filtered settings.}
\label{tab:resultsWN}
\begin{tabular*}{\linewidth}{@{\extracolsep{\fill}}lcccc}
\hline
Model&Mean Rank&Filtered Mean Rank&Hits@1&Filtered Hits@1\\
\hline
KGML ($\alpha = 1.0$) \cite{onuki2019relation} & - &-& 0.73 & 0.94\\
KG-BERT \cite{yao2019kg}   & \textbf{1.02}   &\textbf{1.01}&\textbf{0.98}   & \textbf{0.99}\\
RPLLM (Ours)    &\textbf{1.02}&\textbf{1.01}&\textbf{0.98}& 0.98\\
\hline
\end{tabular*}
\end{table}

\subsection{Main Results}
Table \ref{tab:results} and Table \ref{tab:resultsWN} report our evaluation results for the FreeBase and WordNet KGs respectively. In FreeBase, our model outperforms the best scores in the MR and Hit@1 metrics except the filtered Hits@1 for KG-BERT and KGML ($\alpha = 0.0$), where our model's score is 1\% less. We attribute the filtered Hits@1 score to two main reasons. First, the evaluation procedure in the filtered settings. Precisely, the inclusion of the training and validation triples in the in the evaluation procedure, which boosts the model's scores. Second, the extended training epochs in KG-BERT, which is 20 as suggested by the authors whereas our model is trained for 10 epochs only.

In the WordNet KG evaluation, we believe that our model could not achieve superiority due to two reasons. First, the size of the relations set, which is way smaller than FreeBase. Second, the nature of the dataset, which is a lexical KG for English.

\begin{table}[t]
\centering
\caption{The results in the inductive and the transductive settings on the FreeBase dataset.}
\label{tab:inductive}
\begin{tabular*}{\linewidth}{@{\extracolsep{\fill}}lcc}
\hline
Metric & RPLLM (Inductive) & RPLLM (Transductive) \\
\hline
Mean Rank & 2.58 & 1.49 \\
Filtered Mean Rank & 2.33 & 1.16 \\
Mean Reciprocal Rank & 0.80 & 0.85 \\
Filtered Mean Reciprocal Rank & 0.89 & 0.97 \\
Hits@1 & 0.67 & 0.74 \\
Filtered Hits@1 & 0.79 & 0.95 \\
Hits@5 & 0.96 & 0.99 \\
Filtered Hits@5 & 0.96 & 1.00 \\
\hline
\end{tabular*}
\end{table}

\subsection{Inductive Settings}
To showcase the efficacy of LLMs and text-based RP models in the inductive Settings, that is when predicting relations for text that was not seen during the model training, we evaluate the model on a split of the FreeBase KG that has testing entities not included in the validation and training parts. Additionally, the validation part includes entities not matched with any entity in the training part. We generate these splits based on random 10\% of the original testing triples of FreeBase.

Table \ref{tab:inductive} shows the scores on the same metrics used in the previous results and additional two metrics, that are the MRR and Hits@5, which are described in Section \ref{sec:metrics}. Considerably, the results show inductive settings performance close to the transductive settings. For instance, the MRR score of our model in the transductive settings is only better with a margin of 5\%.


\subsection{Failure Analysis}\label{sec:failure}
Table \ref{tab:failure} reports our model's three lowest test predictions. The three records were sorted the worst based on the rank of the ground truth relation in the predictions. We reason the low rank in the three observations by two causes. First, the limited number of training triples that have the same relations. In other words, the model was not trained enough on the three relations. For instance, the relation \textit{Appointed\_by} in triple \#1 was seen one time in the training part of the dataset. Second, the entity ambiguity problem, that is when the node name represents more than one entity. For example, in triple \#3, it is obscure to identify the entity \textit{Venus} as the solar planet or the roman goddess.

\begin{table}[t]
\centering
\caption{Failure analysis in RPLLM's lowest three test predictions.}
\label{tab:failure}
\begin{tabular*}{\linewidth}{@{\extracolsep{\fill}}lllll}
\hline
ID&Rank&Head Node&Relation&Tail Node\\
\hline
1&944	&	Hillary Rodham Clinton	&\begin{tabular}{l} /people/appointee/position.\\/people/appointment/appointed\_by \end{tabular}&Barack Obama\\
2&791	&	Sun	&/astronomy/celestial\_object/category	&Star\\
3&585&	Venus&	/user/radiusrs/default\_domain/astrology/related\_topics	&Art\\
\hline
\end{tabular*}
\end{table}

\section{Conclusion}
We introduce RPLLM, a utilization of large language models for the relation prediction task. The model is designed to complete knowledge graphs by predicting plausible relations for a given pair of node entities. RPLLM leverages Llama 2, a new and powerful large language model. Our experiments on a well-known benchmark demonstrate that RPLLM surpasses the best scores achieved in the relation prediction task.

\section{Limitations}
While LLMs excel in understanding text sequences, they still require substantial computational resources to operate effectively. As a result, our experiments were constrained to running Llama 2 with 7 billion parameters. However, with additional GPU capabilities, we could leverage the 70 billion parameter version.

Evaluating KGC models could be further enhanced by utilizing unified metrics. For instance, instead of the mean rank, several models employ the mean reciprocal rank. Additionally, models may use varying values of N in the Hits@N metric, such as 3, 5, and 10.

\section{Future Work}
Future KGC research could address the entity ambiguity problem. We demonstrated how this issue led to low rankings in our model's evaluation in Section \ref{sec:failure}. Specifically, future endeavors could explore leveraging the relation text in the training triples, which was not considered in our implementation. Moreover, future efforts could effectively utilize entity descriptions to improve results with minimal additional computational load. Additionally, a promising research direction could enrich the literature by introducing a new evaluation knowledge graph featuring more sophisticated relation prediction scenarios. We claim that current models, including ours, exhibit state-of-the-art performance, evidenced by consistently high reported prediction ranks and small improvements from new studies.

Future research could utilize our work to achieve improved results through additional fine-tuning of the model. Moreover, our implementation could be adapted for the entity prediction task, presenting a novel contribution. Additionally, a potential research direction could explore the application of negative sampling techniques for the RP task, as discussed in Section \ref{sec:experimental setup}.

Finally, while our model is capable of operating in entity inductive settings, it is not designed to handle relation inductive settings. In other words, it cannot predict relations that were not encountered during training. Therefore, future work could address this problem.


\bibliographystyle{ACM-Reference-Format}
\bibliography{references}

\end{document}